\documentclass{article}

\usepackage[pdftex]{graphicx,color}
\usepackage{amsmath}
\usepackage{amssymb}
\usepackage{graphicx}
\usepackage{color}
\usepackage{url}
\usepackage{amsfonts, amscd, amsthm}
\usepackage{algorithm}
\usepackage{algorithmic}

\renewcommand{\mathbf}{\boldsymbol}

\newcommand{\mb}{\mathbf}

\newif\ifincludeintroduction
\newif\ifincludefigures
\includefiguresfalse
\newif\ifincludenumerics

\numberwithin{equation}{section}

\newcommand{\minimize}[2]{\text{minimize} \quad #1 \quad \text{subj. to} \quad #2}

\begin{document}

\title{\bf Sparsity and Robustness in Face Recognition \\ \vspace{3mm}
\large A tutorial on how to apply the models and tools correctly}
\author{John Wright, Arvind Ganesh, Allen Yang, Zihan Zhou, and Yi Ma}
%\date{October 18, 2011}
\date{}
\maketitle

\noindent{\bf Background.} This note concerns the use of techniques for sparse signal representation and sparse error correction for automatic face recognition. Much of the recent interest in these techniques comes from the paper \cite{WrightJ2009-PAMI}, which showed how, under certain technical conditions, one could cast the face recognition problem as one of seeking a sparse representation of a given input face image in terms of a ``dictionary'' of training images and images of individual pixels. To be more precise, the method of \cite{WrightJ2009-PAMI} assumes access to a sufficient number of well-aligned training images of each of the $k$ subjects. These images are stacked as the columns of matrices $\mb A_1, \dots, \mb A_k$. Given a new test image $\mb y$, also well aligned, but possibly subject to illumination variation or occlusion, the method of \cite{WrightJ2009-PAMI} seeks to represent $\mb y$ as a sparse linear combination of the database as whole. Writing $\mb A = [ \mb A_1 \mid \dots \mid \mb A_k ]$, this approach solves 
\[
\minimize{\| \mb x \|_1+\| \mb e \|_1}{\mb A \mb x + \mb e = \mb y}. 
\]
If we let $\mb x_j$ denote the subvector of $\mb x$ corresponding to images of subject $j$, \cite{WrightJ2009-PAMI} assigns as the identity of the test image $\mb y$ the index whose sparse coefficients minimize the residual:
\[
\hat{i} = \arg \min_i \;\| \mb y - \mb A_i \mb x_i - \mb e \|_2. 
\]
This approach demonstrated successful results in laboratory settings (fixed pose, varying illumination, moderate occlusion) in \cite{WrightJ2009-PAMI}, and was extended to more realistic settings (involving moderate pose and misalignemnt) in \cite{Wagner2011-PAMI}. For the sake of clarity, we repeat the above algorithm below. 
\begin{equation} \label{eqn:src}
\text{(SRC)}\left\{\begin{array}{l} \minimize{\| \mb x \|_1+\| \mb e \|_1}{\mb A \mb x + \mb e = \mb y}, \\
\hat{i} = \arg \min_i \;\| \mb y - \mb A_i \mb x_i - \mb e \|_2. 
\end{array} \right. 
\end{equation}
We label this algorithm SRC (sparse representation-based classification), following the naming convention of \cite{WrightJ2009-PAMI}.

A recent paper of Shi and collaborators \cite{ShiQ2011-CVPR} raises a number of criticisms of this approach. In particular, \cite{ShiQ2011-CVPR} suggests that (a) linear representations of the test image $\mb y$ in terms of training images $\mb A_1 \dots \mb A_k$ are not well-founded and (b) that the $\ell^1$-minimization in \eqref{eqn:src} can be replaced with a solution that minimizes the $\ell^2$ residual. In this note, we briefly discuss the analytical and empirical justifications for the method of \cite{WrightJ2009-PAMI}, as well as the implications of the criticisms of \cite{ShiQ2011-CVPR} for robust face recognition. We hope that discussing the discrepancy between the two papers within the context of a richer set of related results will provide a useful tutorial for readers who are new to these concepts and tools, helping to understand their strengths and limitations, and to apply them correctly. 

\section{Linear Models for Face Recognition with Varying Illumination}

The method of \cite{WrightJ2009-PAMI} is based on low-dimensional linear models for illumination variation in face recognition. Namely, the paper assumes that if we have observed a sufficient number of well-aligned training samples $\mb a_1 \dots \mb a_n$ of a given subject $j$, then given a new test image $\mb y$ of the same subject, we can write 
\begin{equation} \label{eqn:model}
\mb y \;\approx\; [ \mb a_1 \mid \dots \mid \mb a_n ] \, \mb x \;\doteq\; \mb A_j \mb x,
\end{equation}
where $\mb x$ is a vector of coefficients. This low-dimensional linear approximation is motivated by theoretical results \cite{Basri2003-PAMI,Frolova2004-ECCV,Ramamoorthi2002-PAMI} showing that well-aligned images of a convex, Lambertian object lie near a low-dimensional linear subspace of the high-dimensional image space. These results were themselves motivated by a wealth of previous empirical evidence of effectiveness of linear subspace approximations for illumination variation in face data (see \cite{Hallinan1994-CVPR,Epstein1995,BelhumeurP1998-IJCV,Yuille1999-IJCV,Georghiades2001-PAMI}).

\begin{figure}[h]
\centerline{
\begin{minipage}{3in} 
\includegraphics[width=3in]{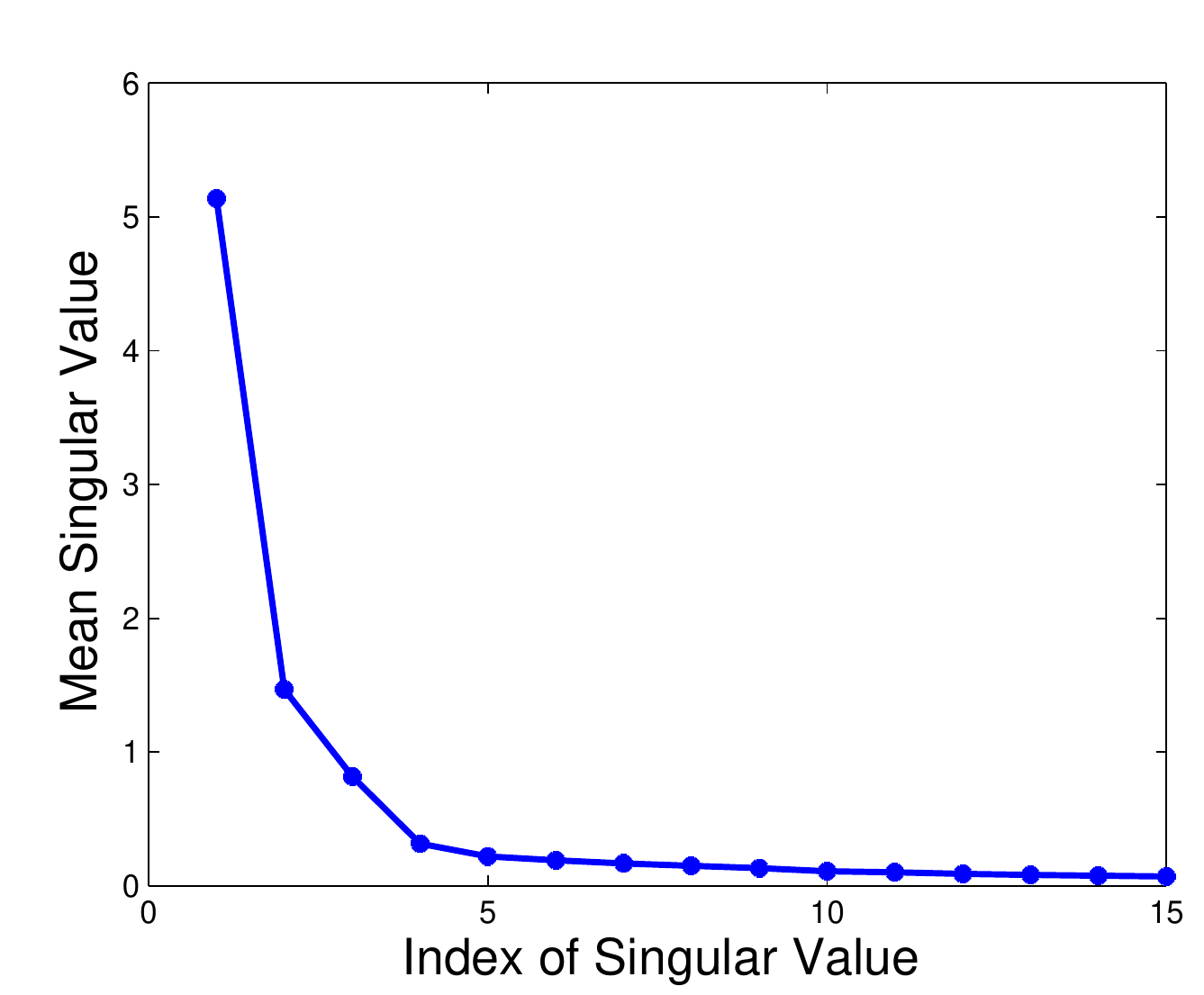}
\centerline{\bf Low rank approximation by SVD}
\end{minipage}
\begin{minipage}{3in}
\includegraphics[width=3in]{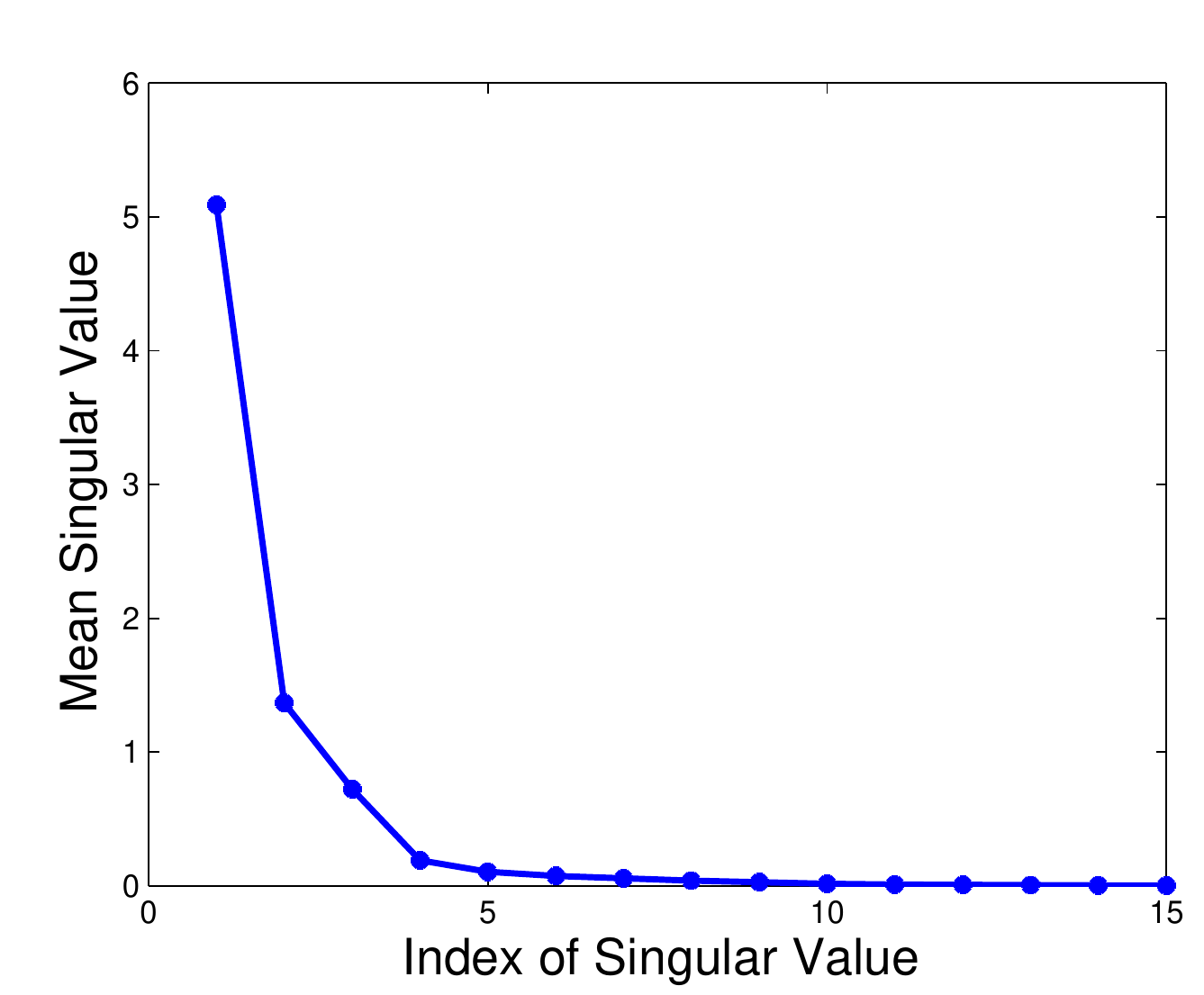}
\centerline{{\bf Low-rank approximation by RPCA}}
\end{minipage}
}
\caption{{\bf Low-dimesional structure in the Extended Yale B database.} We compute low-rank approximations to the images of each subject in the Extended Yale B database, under illumination subsets 1-3. (left) Mean singular values across subjects, when low-rank approximation is computed using singular value decomposition. (right) Mean singular values across subjects, when low-rank approximation is computed robustly using convex optimization. In both cases, the singular values decay; when sparse errors are corrected, the decay is more pronounced. } \label{fig:yaleB}
\end{figure}

To see this phenomenon in the data used in \cite{WrightJ2009-PAMI}, we take Subsets 1-3 of the Extended Yale B database (as used in the experiments by \cite{WrightJ2009-PAMI}). We compute the singular value decomposition of each subject's images. Figure \ref{fig:yaleB} (left) plots the mean of each singular value, across all 38 subjects. We observe that most of the energy is concentrated in the first few singular values. 

Of course, some care is necessary in using these observations to construct algorithms. The following physical phenomena break the low-dimensional linear model:
\begin{itemize}
\item {\bf Specularities and cast shadows} break the assumptions of the low-dimensional linear model. These phenomena are spatially localized, and can be treated as large-magnitude, sparse errors. 
\item {\bf Occlusion} also introduces large-magnitude, sparse errors.
\item {\bf Pose variations and misalignment} introduce highly nonlinear transformations of domain, which break the low-dimensional linear model.
\end{itemize}
Specularities, cast shadows and moderate occlusion can be handled using techniques from sparse error correction. Indeed, using the ``Robust PCA'' technique of \cite{Candes2011-JACM} to remove sparse errors due to cast shadows and specularities, we obtain Figure \ref{fig:yaleB} (right). Once violations of the linear model are corrected, the singular values decay more quickly. Indeed, only the first 9 singular values are significant, corroborating theoretical results of Basri, Ramamoorthi and collaborators. 

The work of \cite{WrightJ2009-PAMI} assumed access to well-aligned training images, with sufficient illuminations to accurately approximate new input images. Whether this assumption holds in practice depends strongly on the scenario. In extreme examples, when only a single training image per subject is available, it will clearly be violated. In applications to security and access control, this assumption can be met: \cite{Wagner2011-PAMI} discusses how to collect sufficient training data for a single subject, and how to deal with misalignment in the test image. Less controlled training data (for example, subject to misalignment) can be dealt with using similar techniques \cite{Peng2011}. 

\begin{figure}[h]
\centerline{
\begin{minipage}{3in} 
\includegraphics[width=3in]{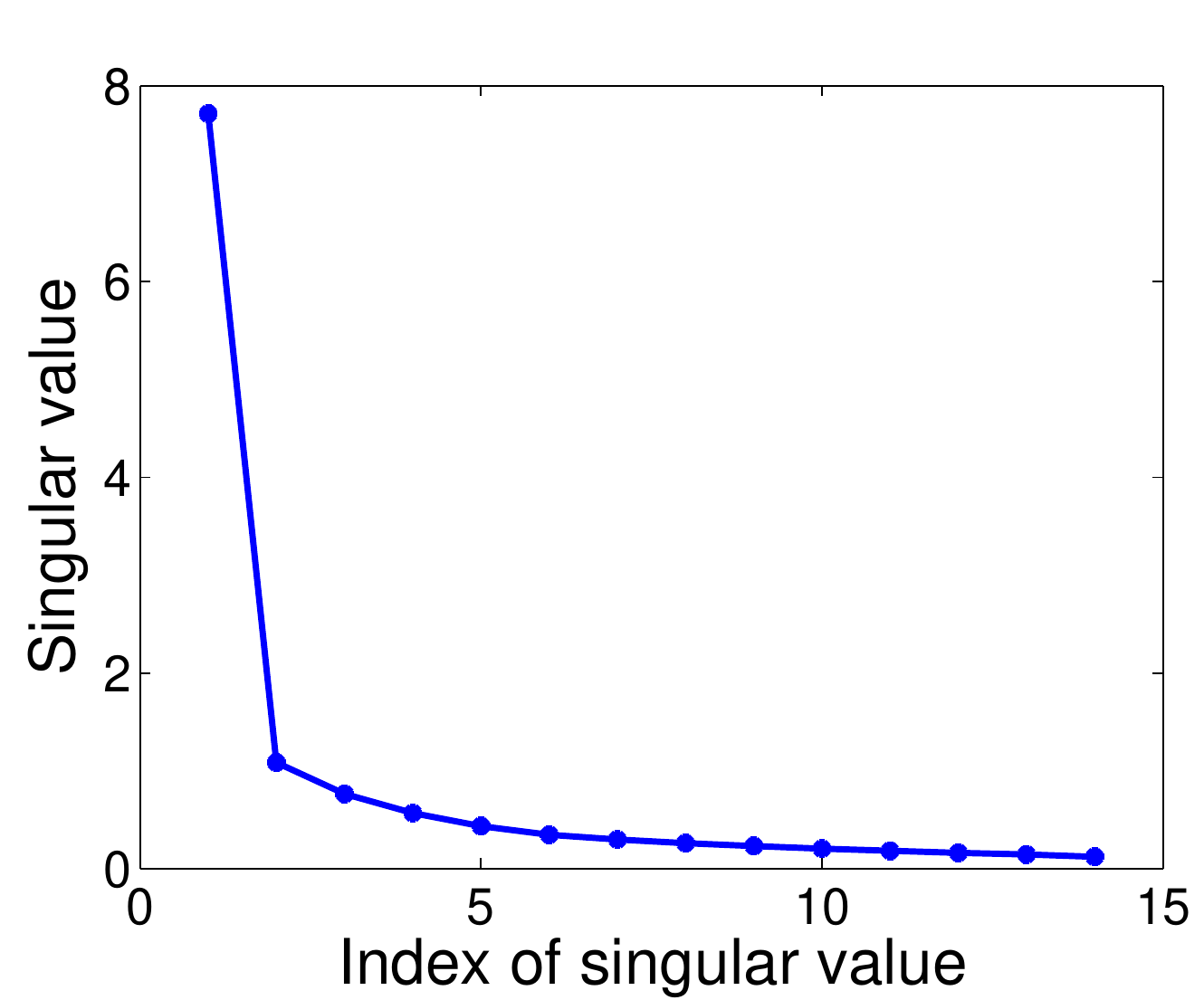}
\centerline{\bf Low rank approximation by SVD}
\end{minipage}
\begin{minipage}{3in}
\includegraphics[width=3in]{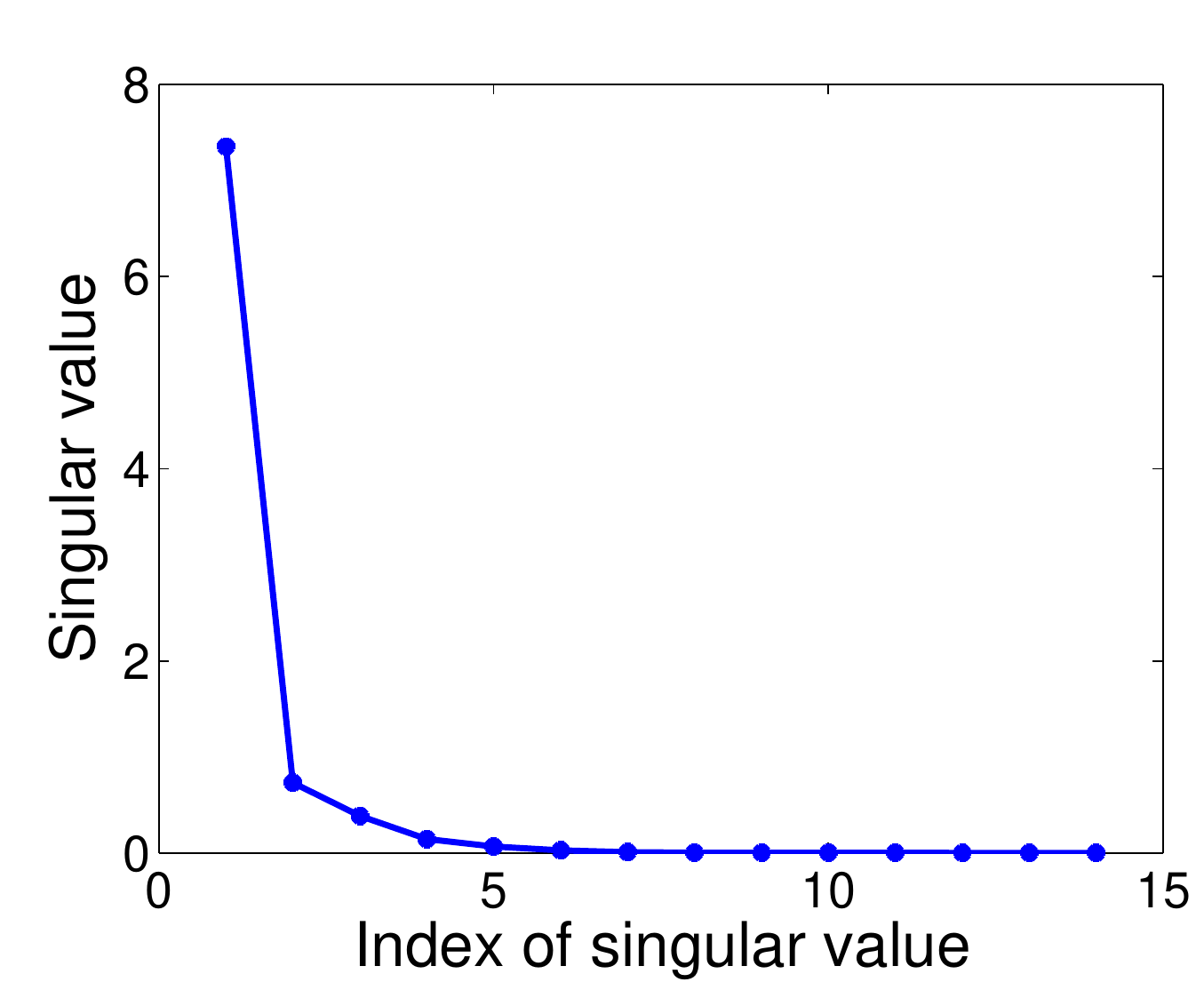}
\centerline{{\bf Low-rank approximation by RPCA}}
\end{minipage}
}
\caption{{\bf Low-dimensional structure in the AR database.} We compute low-rank approximations to the images of each subject in the AR database, using images with varying illumination and expression (14 images per subject). (left) Mean singular values across subjects, when low-rank approximation is computed using singular value decomposition. (right) Mean singular values across subjects, when low-rank approximation is computed robustly using convex optimization. Again, in both cases, the singular values decay; when sparse errors are corrected, the decay is more pronounced.} \label{fig:AR}
\end{figure}

The above experiments use the Extended Yale B face database, which was constructed to investigate illumination variations in face recognition. However, similar results can be obtained on other datasets. We demonstrate this using the AR database, which was also used in the experiments of \cite{WrightJ2009-PAMI}. We take the cropped images from this database, with varying expression and illumination. There are a total of 14 images per subject. Figure \ref{fig:AR} plots the resulting singular values obtained via SVD (left) and with a robust low-rank approximation (right). One can clearly observe low-rank structure\footnote{In fact, when the low-rank approximation is computed robustly, its numerical rank always lies in the range of $6-9$. However, this number is less important than the singular values themselves, which decay quickly.}. However, this structure does not necessarily arise from the Lambertian model -- the number of distinct illuminations may not be sufficient, and some subjects' images have significant saturation. Rather, the low-rank structure in the AR database arises from the fact that conditions are repeated over time. 

%We use the technique of \cite{Peng2011} to align them and correct sparse errors\footnote{This is done in a fully automatic manner, using output of a face detector as initialization.}. Alignment is necessary here because the subjects' head position changes between sessions. 

\paragraph{Comments on the ``assumption test" by Shi et.\ al.} \cite{ShiQ2011-CVPR} report the following experimental result: all of the cropped images from {\em all subjects} of the AR database are stacked as columns of a large matrix $\mb A$. The singular values of $\mb A$ are computed. The singular values of this matrix are peaked in the first few entries, but have a heavy tail. Because of this, \cite{ShiQ2011-CVPR}  conclude that images of {\em a single subject} in AR do not exhibit low-dimensional linear structure. Their observation does not imply this conclusion, for at least two reasons: 

\begin{itemize}

\item First, low-dimensional linear structure is expected to occur within the images of a single subject. The distribution of singular values of a dataset of many subjects as a whole depends not only on the physical properties of each subject's images, but on the distribution of face shapes and reflectances across the population of interest. Investigating properties of the singular values of the database as a whole is a questionable way to test hypotheses about the numerical rank or spectrum of a single subject's images. This is especially the case when each subject's images are not perfectly rank deficient, but rather approximated by a low-dimensional subspace (as is implied by \cite{Basri2003-PAMI}): the overall spectrum of the matrix will depend significantly on the relative orientation of all the subspaces.\footnote{Indeed, \cite{ShiQ2011-CVPR} observe a distribution of singular values across all the subjects that resembles the singular values of a Gaussian matrix. This is reminiscent of \cite{WrightJ2010-IT}, in which the the uncorrupted training images of many subjects are modeled as small Gaussian deviations about a common mean. The implications of such a model for error correction are rigorously analyzed in \cite{WrightJ2010-IT}. It should also be noted that the values of the plotted singular values in \cite{ShiQ2011-CVPR} are not, as suggested, the singular values of a standard Gaussian matrix of the same size as the test database -- they are the singular values of a smaller, {\em square} Gaussian random matrix, and hence do not reflect the noise floor in the AR database.}

\item Second, the images used in the experiment of Shi et.\ al.\ include occlusions, and may not be precisely aligned at the pixel level. Both of these effects are known to break low-dimensional linear models. Indeed, above, we saw that if we restrict our attention to training images that do not have occlusion (as in \cite{WrightJ2009-PAMI}) and compute robustly, low-dimensional linear structure becomes evident. 

%\item  Finally, the plotted singular values in \cite{ShiQ2011-CVPR} are not, as suggested, the singular values of a standard Gaussian matrix of the same size as the test database. They are the singular values of a {\em square} Gaussian random matrix, which is smaller than the data matrix. Their values may not reflect the noise floor in the AR database. As already suggested by \cite{WrightJ2010-IT}, well before \cite{ShiQ2011-CVPR}, a possible model for (un-corrupted) training images of many subjects is a random bouquet model -- a random Gaussian matrix with a large mean and small variance for the columns. The properties of such a model and its implications for (even dense) error correction has been thoroughly analyzed and proven in \cite{WrightJ2010-IT}. 
\end{itemize}

\section{Robustness, $\ell^1$ and the $\ell^2$ Alternatives}

In the previous section, we saw that images of the same face under varying illumination could be well-represented using a low-dimensional linear subspace, provided they were well-aligned and provided one could correct gross errors due to cast shadows and specularities. These errors are prevalent in real face images, as are additional violations of the linear model due to occlusion. Like specular highlights, the error incurred by occlusion can be large in magnitude, but is confined to only a fraction of the image pixels -- it is {\em sparse} in the pixel domain. In \cite{WrightJ2009-PAMI}, this effect is modeled using an additive error $\mb e$. If the only prior information we have about $\mb e$ is that it is sparse, then the appropriate optimization problem becomes 
\begin{equation} \label{eqn:combined}
\minimize{\| \mb x \|_1 + \| \mb e \|_1}{\mb y = \mb A \mb x + \mb e}.
\end{equation}
Clearly, any robustness derived from the solution to this optimization problem is due to the presence of the sparse error term, and the minimization of the $\ell^1$ norm of $\mb e$. Indeed, based on theoretical results in sparse error correction, we should expect that the above $\ell^1$ minimization problem will successfully correct the errors $\mb e$ provided the number of errors (corrupted, occluded or specular pixels) is not too large. For certain classes of matrices $\mb A$ one can identify sharp thresholds on the number of errors, below which $\ell^1$ minimization performs perfectly, and beyond which it breaks down. In contrast, minimization of the $\ell^2$ residual, say $\min \| \mb y - \mb A \mb x\|_2$ does not have this property. 

The paper of \cite{ShiQ2011-CVPR} suggests that the use of the $\ell^1$ norm in \eqref{eqn:combined} is unnecessary, and proposes two algorithms. The first solves
\begin{equation} \label{eqn:shi1}
\text{($\ell^2$-1)}\quad \left\{\begin{array}{l} \text{minimize} \quad \| \mb y - \mb A \mb x \|_2, \\
\hat{i} = \arg \min_i \;\| \mb y - \mb A_i \mb x_i\|_2. 
\end{array} \right. 
\end{equation}
This approach is not expected to be robust to errors or occlusion. For faces occluded with sunglasses and scarves (as in the AR Face Database), \cite{ShiQ2011-CVPR} suggests an extension 
\begin{equation} \label{eqn:shi2}
\text{($\ell^2$-2)}\quad \left\{\begin{array}{l} \text{minimize} \quad \| \mb y - \mb A \mb x - \mb W \mb v \|_2, \\
\hat{i} = \arg \min_i \;\| \mb y - \mb A_i \mb x_i  \|_2. 
\end{array} \right.
\end{equation}
where $\mb W$ is a tall matrix whose columns are chosen as blocks that may well-represent occlusions of this nature. 

In trying to understand the strengths and working conditions of these proposals several questions arise. First, do the approaches (SRC), ($\ell^2$-1)  and ($\ell^2$-2) provide robustness to general pixel-sparse errors? We test this using settings and data {\em identical} to those in \cite{WrightJ2009-PAMI}, in which the Extended Yale B database subsets I and II are used for training, and subset III is used for testing. Varying fractions of random pixel corruption are added, from 0\% to 90\%. Table \ref{tab:tab1} shows the resulting recognition rates for the three algorithms. The $\ell^1$ minimization \eqref{eqn:combined} is robust to up to 60-70\% arbitrary random errors. In contrast, both methods based on $\ell^2$ minimization break down much more quickly. We note that this result is expected from theory: \cite{WrightJ2010-IT} provides results in this direction.\footnote{To be precise, results in \cite{WrightJ2010-IT} suggest, but do not prove, that $\ell^1$ will succeed at correcting large fractions of errors in this situation. The rigorous theoretical results of \cite{WrightJ2010-IT} pertain to a specific stochastic model for $\mb A$.} To be clear, the goal of this experiment is not to assert that the $\ell^1$ norm is ``better'' or ``worse'' than $\ell^2$ in some general sense -- simply to show that $\ell^1$ provides robustness to general sparse errors, whereas the two approaches \eqref{eqn:shi1}-\eqref{eqn:shi2} do not.  There are situations in which it is correct (optimal, in fact) to minimize the $\ell^2$ norm -- when the error is expected to be dense, and in particular, if it follows an iid Gaussian prior. However, for sparse errors, $\ell^1$ has well-justified and thoroughly documented advantages. 

\begin{table}[!ht]
\centering
\begin{tabular}{|c|c|c|c|}
\hline
& \multicolumn{3}{|c|}{Recognition rate (\%)} \\
{\bf \% corrupted pixels} & {\bf SRC} & {\bf $\ell^2$-1} & {\bf $\ell^2$-2} \\
\hline
\hline
0 & 100 & 100 & 100 \\
\hline
10 & 100 & 100 & 100 \\
\hline
20 & {\bf 100} & 99.78 & 99.78 \\
\hline
30 & {\bf 100} & 99.56 & 99.34 \\
\hline
40 & {\bf 100} & 96.25 & 96.03 \\
\hline
50 & {\bf 100} & 83.44 & 81.23 \\
\hline
60 & {\bf 99.3} & 59.38 & 59.94 \\
\hline
70 & {\bf 90.7} & 38.85 & 40.18 \\
\hline
80 & {\bf 37.5} & 15.89 & 15.23 \\
\hline
90 & 7.1 & 8.17 & {\bf 7.28}\\
\hline
\end{tabular}
\caption{{\bf Extended Yale B database with random corruption.} Subsets 1 and 2 are used as training and Subset 3 as testing. The best recognition rates are in bold face. SRC ($\ell^1$) performs robustly up to about 60\% corruption, and then breaks down. Alternatives are significantly less robust.}
\label{tab:tab1}
\end{table}

Of course, real occlusions in images are very different in nature for the random corruptions considered above -- occlusions are often spatially contiguous, for example. Hence, we next ask to what extent the three methods provide robustness against general spatially contiguous errors. We investigate this using random synthetic block occlusions {\em exactly the same} as in \cite{WrightJ2009-PAMI}. The results are reported in Table \ref{tab:tab2}.

\begin{table}[!ht]
\centering
\begin{tabular}{|c|c|c|c|}
\hline
& \multicolumn{3}{|c|}{Recognition rate (\%)} \\
{\bf \% occluded pixels} & {\bf SRC} & {\bf $\ell^2$-1} & {\bf $\ell^2$-2} \\
\hline
\hline
10 & {\bf 100} & 99.56 & 99.78 \\ \hline
20 & {\bf 99.8} & 95.36 & 97.79 \\ \hline
30 & {\bf 98.5} & 87.42 & 92.72 \\ \hline
40 & {\bf 90.3} & 76.82 & 82.56 \\ \hline
50 & 65.3 & 60.93 & {\bf 66.22}\\ \hline
\end{tabular}
\caption{{\bf Extended Yale B with block occlusions.} Subsets 1 and 2 are used as training, Subset 3 as testing. The best recognition rates are in bold face. SRC $\ell^1$ minimization performs quite well upto a breakdown point near 30\% occluded pixels, then breaks down. The two alternatives based on $\ell^2$ norm minimization degrade more rapidly as the frraction of occlusion increases.}
\label{tab:tab2}
\end{table}

Notice that again, $\ell^1$ minimization performs more robustly than either of the $\ell^2$ alternatives. As in the previous experiment, the good performance compared to ($\ell^2$-1) is expected (indeed, \cite{ShiQ2011-CVPR} do not assert that ($\ell^2$-1) is robust against error). The good performance compared to ($\ell^2$-2) is also expected, as the basis $\mb W$ is designed for certain specific errors (incurred by sunglasses and scarves). It is also important to note that the breakdown point for $\ell^1$ with spatially coherent errors is lower than for random errors ($\approx 30\%$ compared to $\approx 60\%$). Again, this is expected -- the theory of $\ell^1$ minimization suggests the existence of a worst case breakdown point (the strong threshold), which is lower than the breakdown point for randomly supported solutions (the weak threshold). For spatially coherent errors, we should not expect $\ell^1$ minimization to succeed beyond this threshold of 30\%. Nevertheless, if one could incorporate the spatial continuity prior of the error support in a principled manner, one could expect to see $\ell^1$ minimization to tolerate more than $60\%$ errors, as investigated further in \cite{ZhouZ2009-ICCV}, well before the work of \cite{ShiQ2011-CVPR}.

Finally, to what extent do the three methods provide robustness to the specific real occlusions encountered in the AR database? Here, we should distinguish between two cases -- occlusion by sunglasses and occlusion by scarves. Sunglasses fall closer to the aforementioned threhold, whereas scarves significantly violate it, covering over 40\% of the face. Table \ref{tab:tab3} shows the results of the three methods for these types of occlusion, at the same image resolution used in \cite{WrightJ2009-PAMI} ($80 \times 60$).\footnote{The basis images used in forming the matrix $\mb W$ are transformed to this size using Matlab's imresize command.} 

\begin{table}[!ht]
\centering
\begin{tabular}{|c|c|c|c|c|}
\hline
& \multicolumn{4}{|c|}{Recognition rate (\%)} \\
{\bf Occlusion type} & {\bf SRC} & {\bf $\ell^2$-1} & {\bf $\ell^2$-2} & \cite{ZhouZ2009-ICCV}\\
\hline
\hline
Sunglasses & {87} & 59.5 & 83 & 99--{\bf 100}\\ \hline
Scarf & 59.5 & {85} & 82.5 & 97--{\bf 97.5} \\ \hline
\end{tabular}
\caption{{\bf AR database, with the data and settings of \cite{WrightJ2009-PAMI}.} SRC outperforms $\ell^2$ alternatives for sunglasses, but does not handle occlusion by scarves well, as it falls beyond the breakdown point for contiguous occlusion.}
\label{tab:tab3}
\end{table}

From Table 3, one can see that none of the three methods is particularly satisfactory in its performance. For sunglasses, $\ell^1$ norm minimization outperforms both $\ell^2$ alternatives. Scarves fall beyond the breakdown point of $\ell^1$ minimization, and SRC's performance is, as expected, unsatisfactory. The performance of ($\ell^2$-2) for this case is better, although none of  the methods offers the strong robustness that we saw above for the Yale dataset. This is the case despite the fact that the basis $\mb W$ in ($\ell^2$-2) was chosen specifically for real occlusions. 

There may be several reasons for the above unsatisfactory results on the AR database: 1. Unlike the Yale database, the AR database does not have many illuminations and images are not particularly well aligned either -- all may compromise the validity of the linear model assumed. 2. None of the models and solutions is particularly effective in exploiting the spatial continuity of the large error supports like sunglasses or scarfs. 

A much more effective way of harnessing the spatial continuity of the error supports was investigated in \cite{ZhouZ2009-ICCV}, where $\ell^1$ minimization, together with a Markov random field model for the errors, can achieve nearly 100\% recognition rates for sunglasses and scarfs with exactly the same setting (trainings, resolution) as above experiments on the AR database.

\section{Comparison on the AR Database with Full-Resolution Images}

Readers versed in the literature on error correction (or $\ell^1$-minimization) will recognize that its good performance is largely a {\em high-dimensional} phenomenon. In the previous examples, it is natural to wonder what lost when we run the methods at lower resolution ($80 \times 60$). In this section, we compare the three methods at the native resolution $165 \times 120$ of the cropped AR database. This is possible thanks to scalable methods for $\ell^1$ minimization \cite{YangA2010-techreport}. 

We use a training set consisting of 5 images per subject -- four neutral expressions under different lighting, and one anger expression, which is close to neutral, all taken under with the same expression. From the training set of \cite{WrightJ2009-PAMI}, we removed three images with large expression (smile and scream), as these effects violate the low-dimensional linear model. In the cropped AR database, for each person, the training set consists of images 1, 3, 5, 6 and 7. The other 8 images per person from Session 1 were used for testing. Table \ref{tab:tab4} lists the recognition rates for each category of test image. Note that there are 100 test images (1 per person) in each category. For these experiments, we use an Augmented Lagrange Multiplier  (ALM)  algorithm to solve the $\ell^1$ minimization problem (see \cite{YangA2010-techreport} for more details). Our Matlab implementation requires on average 259 seconds per test image, when run on a MacPro with two 2.66 GHz Dual-Core Intel Xenon processers and 4GB of memory.\footnote{With 8 images per subject, as in \cite{WrightJ2009-PAMI}, this same approach requires 378 seconds per test image.} We would like to point out that there is scope for improvement in the speed of our implementation. But since this is not the focus of our discussion here, we have used a simple, straightforward version of the ALM algorithm that is accurate but not necessarily very efficient.  In addition, we have used a single-core implementation. The ALM algorithm is very easily amenable to parallelization, and this could greatly reduce the running time, especially when we have a large number of subjects in the database.

\begin{table}[!ht]
\centering
\begin{tabular}{|c|c|c|c|}
\hline
& \multicolumn{3}{|c|}{Recognition rate (\%)} \\
{\bf Test Image category} & {\bf SRC} & {\bf $\ell^2$-1} & {\bf $\ell^2$-2} \\
\hline
\hline
Smile & {\bf 100} & 97 & 95 \\ \hline
Scream & {\bf 88} & 60 & 59 \\ \hline
Sunglass (neutral lighting) & {\bf 88} & 68 & {\bf 88} \\ \hline
Sunglass (lighting 1) & 75 & 63 & {\bf 88} \\ \hline
Sunglass (lighting 2) & {\bf 90} & 69 & 84 \\ \hline
Scarf (neutral lighting) & 65 & 66 & {\bf 76} \\ \hline
Scarf (lighting 1) & {\bf 66} & 63 & 65 \\ \hline
Scarf (lighting 2) & {\bf 68} & 62 & 67 \\ \hline
{\bf Overall} & {\bf 80} & 68.5 & 77.75 \\ \hline
\end{tabular}
\caption{{\bf AR database with 5 training images per person and full resolution}. The best recognition rates are in bold face.}
\label{tab:tab4}
\end{table}

From the above experiment, we can see that when the three approaches are compared with images of the same resolution, the results differ significantly from those of \cite{ShiQ2011-CVPR}. We will explain this discrepency in the next section. 

On the other hand, we observe that none of the methods performs in a completely satisfactory manner on images with large occlusion -- in particular, images with scarves. This is expected from our experiments in the previous section. Can strong robustness (like that exhibited by SRC with $\le 60\%$ random errors or $\le 30\%$ contiguous errors) be achieved here? It certainly seems plausible, since neither SRC nor ($\ell^2$-1) take advantage of spatial coherence of real occlusions. ($\ell^2$-2) does take advantage of spatial properties of real occlusions, through the construction of the matrix $\mb W$, but it is not clear if or how one can construct a $\mb W$ that is guaranteed to work for all practical cases. 

In \cite{WrightJ2009-PAMI}, $\ell^1$-norm minimization together with a partitioning heuristic is shown to produce much improved recognition rates on the particular cases encountered in AR (97.5\% for sunglasses and 93.5\% for scarfs). However, the choice  of partition is somewhat arbitrary, and this heuristic suffers from many of the same conceptual drawbacks as the introduction of a specific basis $\mb W$. Several groups have studied more principled schemes for exploiting prior information on the spatial layout of sparse signals or errors (see \cite{ZhouZ2009-ICCV} and the references therein). For instance, one could expect that the modified $\ell^1$ minimization method given in \cite{ZhouZ2009-ICCV} would work equally well under the setting (training and resolution) of the above experiments as it did under the setting in the previous section (see Table \ref{tab:tab3}).

\section{Face recognition with low-dimensional measurements}

The results in the previous section, and conclusions that one may draw from them, are quite different from those obtained by Shi et.\ al.\ \cite{ShiQ2011-CVPR}. The reasons for this discrepancy are simple: 
\begin{itemize}
\item In \cite{ShiQ2011-CVPR}, the authors did not solve \eqref{eqn:src} to compare with \cite{WrightJ2009-PAMI}. Rather, they solved\footnote{It seems likely that the authors of \cite{ShiQ2011-CVPR} mistakenly solved instead the following problem: $\minimize{\| \mb x \|_1+ \| \mb e' \|_1}{\mb \Phi \mb y = \mb \Phi \mb A \mb x + \mb e' }$. If that was the case, their results would be even more problematic as the projected error $\mb  e' = \Phi \mb e$ is no longer sparse for an arbitrary random projection. In practice, the sparsity of $\mb e'$ can only be ensured if the projection is a simple downsampling.}
\begin{equation} \label{eqn:projected}
\minimize{\| \mb x \|_1+ \| \mb e \|_1}{\mb \Phi \mb y = \mb \Phi( \mb A \mb x + \mb e )},
\end{equation}
where $\mb \Phi$ is a random projection matrix mapping from the $165 \times 120 = 19,800$-dimensional image space into a meager $300$-dimensional feature space. Using these drastically lower (300) dimensional features, they obtain recognition rates of around 40\% for the above $\ell^1$ minimization, which is compared to a 78\% recognition rate obtained with ($\ell^2$-2) on the full  ($19,800$) image dimension.  As we saw in the previous section, when the two methods are compared on a fair footing with the same number of observation dimensions, the conclusions become very different.

\item In Section 5 of \cite{ShiQ2011-CVPR}, there is an additional issue: the training images in $\mb A$ are randomly selected from the AR dataset sessions regardless of  their nature. In particular, the training and test sets could contain images with significant occlusion. This choice is very different from any of the experimental settings in \cite{WrightJ2009-PAMI},\footnote{In \cite{ShiQ2011-CVPR}, the authors claim that they ``form the matrix $\mb A$ in the same manner as \cite{WrightJ2009-PAMI}''. That is simply {\em not true}.} and also different from settings of all of the above experiments. In Section 1, we have already discussed the problems with such a choice and how it differs from the work of \cite{WrightJ2009-PAMI}.

%\item {\bf In \cite{ShiQ2011-CVPR}, to benchmark the performance of SRC using $\ell_1$-min, the authors chose an inferior numerical solver of Orthogonal Matching Pursuit (OMP). The choice is clearly different from the exact $\ell_1$-min solution of Basis Pursuit (BP) used in \cite{WrightJ2009-PAMI} and several subsequent papers. Nevertheless, it is also well known that exact BP algorithms are computationally expensive in high-dimensional space. Therefore, a practitioner who attempts to accelerate the speed of $\ell_1$-min must proceed with caution, as any fast numerical solution must provide guarantee of convergence that the estimated solution closely approximates the ground-truth sparse signal. We refer the reader to our recent work \cite{YangA2010-techreport} for more thorough examination of the accelerated $\ell_1$-min issue.}
\end{itemize}

The main methodological flaw of \cite{ShiQ2011-CVPR} is to compare the performance of the two methods with dramatically different numbers of measurements -- and in a situation that is quite different from what was advocated in \cite{WrightJ2009-PAMI}:
\begin{itemize}
\item It is easy to see that the minimizer in \eqref{eqn:projected} can have at most $d = 300$ nonzero entries -- far less than the cardinality of the occlusion such as sun glasses or scarf. $\ell^1$ minimization will not succeed in this scenario. In fact, both ($\ell^2$-1) and ($\ell^2$-2) also fail when applied with this set of $d=300$ features. Without proper regularization on $\mb x$ (say via the $\ell^1$-norm), ($\ell^2$-1) and ($\ell^2$-2) have infinite many minimizers, and the approach suggested in \cite{ShiQ2011-CVPR} cannot apply.  
 \item \cite{WrightJ2009-PAMI} also investigated empirically the use random projections as features, for images that are not occluded or corrupted! The model is strictly $\mb y = \mb A \mb x$ (or $\mb y = \mb A \mb x + \mb z$, where $\mb z$ is small (Gaussian) noise) -- no gross errors are involved. As the problem of solving for $\mb x$ from $\mb y = \mb A \mb x$ is underdetermined, $\ell^1$ regularization on $\mb x$ becomes necessarily to obtain the correct solution. However, \cite{WrightJ2009-PAMI} does not suggest that a random projection into a lower-dimensional space can improve robustness -- this is provably false. It also does not suggest solving \eqref{eqn:projected} in cases with errors  -- as the results of \cite{ShiQ2011-CVPR} suggest, this does not work particularly well. 
 \item Nevertheless, {\em under very special conditions}, robustness can still be achieved with severely low-dimensional measurements. As investigated in \cite{ZhouZ2009-ICCV}, if the low-dimensional measures are from down-sampling (that respects the spatial continuity of the errors) and the spatial continuity of the error supports is effectively exploited using a Markov random field model, one can achieve nearly 90\% recognition rates for scarfs and sunglasses at the resolution of $13\times 9$ -- only 111 measurements (pixels), far below the 300 (random) measurements used in \cite{ShiQ2011-CVPR}.
 \end{itemize}

\section{Linear models and solutions}
Like face recognition, many other problems in computer vision or pattern recognition can be cast as solving a set of linear equations, $\mb y = \mb A \mb x + \mb e$. Some care is necessary to do this correctly:
\begin{enumerate}
\item The first step is to verify that the linear model $\mb y = \mb A \mb x + \mb e$ is valid, ideally via physical modeling corroborated by numerical experiments. If the training $\mb A$ and the test $\mb y$ are not prepared in a way such a model is valid, two things could happen: 1.\ there might be no solution or no (unique) solution to the equations; 2.\ the solution can be irrelevant to what you want.
\item The second step, based on the properties of the desired $\mb x$ (least energy or entropy) and those of the errors $\mb e$ (dense Gaussian or sparse Laplacian), one needs to choose the correct optimization objective in order to obtain the correct solution. 
\end{enumerate}
There are already four possible combinations of $\ell^1$ and $\ell^2$ norms\footnote{In the literature, many other norms are also being investigated such as the $\ell^{2,1}$ norm for block sparsity etc.}:
\begin{eqnarray*}
&&\minimize{\| \mb x \|_1 + \| \mb e \|_1}{\mb y = \mb A \mb x + \mb e}  \quad \mbox{(least entropy \& error correction)}\\
&&\minimize{\| \mb x \|_2 + \| \mb e \|_1}{\mb y = \mb A \mb x + \mb e}  \quad \mbox{(least energy \& error correction)}\\
&&\minimize{\| \mb x \|_1 + \| \mb e \|_2}{\mb y = \mb A \mb x + \mb e}  \quad \mbox{(sparse regression with noise -- lasso)}\\
&&\minimize{\| \mb x \|_2 + \| \mb e \|_2}{\mb y = \mb A \mb x + \mb e}  \quad \mbox{(least energy with noise)}
\end{eqnarray*}
Ideally, the question should not be which formulation yields better performance on a specific dataset, but rather which assumptions match the setting of the problem, and then whether the adopted regularizer helps find the correct solution under these assumptions. For instance, when $\mb A$ is under-determined, regularization on $\mb x$ with either the $\ell^1$ or the $\ell^2$ norm is necessary to ensure a unique solution. But the solution can be rather different for each norm. If $\mb A$ is over-determined, the choice of regularizer on $\mb x$ is less important or even is unnecessary. Furthermore, be aware that all above programs could fail (to find the correct solution) beyond their range of working conditions. Beyond the range, it becomes necessary to exploit additional structure or information about the signals ($\mb x$ or $\mb e$) such as spatial continuity etc. 

%When using them in practice, {\bf the question should never be about which formulation is better or worse; but it is only about for a specific problem or data, which formulation(s) give the correct solution.}

\bibliographystyle{alpha}
\bibliography{paper}

\end{document}